\documentclass[lettersize,journal]{IEEEtran}
\usepackage{amsmath,amsfonts}
\usepackage{array}
\usepackage[caption=false,font=normalsize,labelfont=sf,textfont=sf]{subfig}
\usepackage{textcomp}
\usepackage{stfloats}
\usepackage{url}
\usepackage{verbatim}
\usepackage{graphicx}
\usepackage{booktabs}
\usepackage{cite}
\usepackage{amsmath}
\usepackage{paralist}

\DeclareMathOperator*{\argmin}{argmin}

\usepackage{color}
\usepackage{url}
\definecolor{indigo}{rgb}{0.0, 0.25, 0.42}
\definecolor{darkorange}{rgb}{1.0, 0.55, 0.0}
\definecolor{darkblue}{rgb}{0.122, 0.435, 0.698}

\usepackage{microtype}
\usepackage{graphicx}
\usepackage{booktabs} 

\usepackage{algorithm}
\usepackage{algpseudocode}

\usepackage{amsmath}
\usepackage{amssymb}
\usepackage{mathtools}
\usepackage{amsthm}
\usepackage{stfloats}

\usepackage{amsmath}
\usepackage{amssymb}
\usepackage{mathtools}
\usepackage{amsthm}
\usepackage{stfloats}
\usepackage{hyperref}
\usepackage[capitalize,noabbrev]{cleveref}

\newtheorem{definition}{Definition}
\newtheorem{remark}{Remark}

\newtheorem{lemma}{Lemma}

\usepackage{titlesec}
\titlespacing*{\section}{0pt}{3pt}{3pt}
\titlespacing*{\subsection}{0pt}{3pt}{3pt} 
\begin{document}

\title{Over-the-Air Fair Federated Learning via Multi-Objective Optimization}

\author{
Authors
\thanks{
The results of this study have been submitted in part to 2025 IEEE International Conference on Acoustics, Speech, and Signal Processing (ICASSP).
}
}
\author{Shayan Mohajer Hamidi,~\IEEEmembership{Student Member,~IEEE}, Ali Bereyhi,~\IEEEmembership{Member,~IEEE}, Saba Asaad, ~\IEEEmembership{Member,~IEEE},\\ and H. Vincent Poor, ~\IEEEmembership{Life Fellow,~IEEE}.
}

\maketitle

\begin{abstract}
In federated learning (FL), heterogeneity among the local dataset distributions of clients can result in unsatisfactory performance for some, leading to an unfair model. To address this challenge, we propose an \textbf{o}ver-\textbf{t}he-\textbf{a}ir \textbf{f}air \textbf{f}ederated \textbf{l}earning algorithm (OTA-FFL), which leverages over-the-air computation to train fair FL models. By formulating FL as a multi-objective minimization problem, we introduce a modified Chebyshev approach to compute adaptive weighting coefficients for gradient aggregation in each communication round. To enable efficient aggregation over the multiple access channel, we derive analytical solutions for the optimal transmit scalars at the clients and the de-noising scalar at the parameter server. Extensive experiments demonstrate the superiority of OTA-FFL in achieving fairness and robust performance compared to existing methods.    
\end{abstract}

\begin{IEEEkeywords}
Federated learning, wireless communications, fairness, multi-objective optimization.
\end{IEEEkeywords}

\section{Introduction}
\label{Sec:Introduction}
\IEEEPARstart{M}{achine} learning (ML) models have traditionally been trained in a centralized manner, where all training data is collected and stored in a central data center or cloud server. However, in many modern applications, sharing sensitive data with remote servers is often infeasible due to privacy concerns. Federated learning (FL) addresses this issue by enabling devices to collaboratively train a global model using their local datasets, coordinated by a parameter server (PS) \cite{mcmahan2017communication}. In this framework, only local model updates are transmitted to the PS, ensuring that raw data remains private. Given that these updates are typically transmitted over wireless channels \cite{10487854}, extensive research has focused on integrating FL into wireless networks \cite{amiri2020machine,yang2020federated}.

To achieve efficiency in wireless FL, over-the-air (OTA) computation has been widely adopted for effective uplink model transmission \cite{amiri2020machine,9810113,9833972,yang2020federated}. By leveraging the waveform superposition property of multiple-access channel (MAC), OTA enables simultaneous model transmission and aggregation. This approach significantly reduces communication latency and conserves uplink communication bandwidth.
A notable challenge in FL, including OTA-FL, stems from the heterogeneity in the distributions of clients' local datasets \cite{10619204}. This heterogeneity can lead to poor performance when the global model is applied to individual clients’ private datasets, resulting in an \textit{unfair} global model \cite{li2019fair}. Specifically, while the average accuracy across clients may be high, clients with data distributions that differ significantly from the majority are more likely to experience degraded model performance. To highlight the importance of designing a \textit{fair} FL algorithm, consider the development of personalized healthcare models based on data from multiple hospitals. In this scenario, it is essential to ensure that the healthcare model is equitable and does not produce biased treatment recommendations for certain patient groups.

In the literature, several studies have sought to train fair models under the assumption that the PS receives noise-free gradients from clients \cite{li2019fair,li2020tilted,hamidi2024distributed}. While these methods achieve fairness in idealized settings, they fall short in practical scenarios where wireless channel imperfections introduce noise into the received local updates \cite{10381881}. To address this, \cite{10381881} proposed a fair FL method that accounts for wireless channel imperfections. However, their approach requires the PS to have direct access to each client’s individual gradient vector, making it incompatible with OTA computation techniques.

To bridge this gap, this paper introduces an over-the-air fair federated learning algorithm (OTA-FFL). Inspired by \cite{hu2022federated,hamidi2024adafed}, OTA-FFL formulates FL as a multi-objective minimization (MoM) problem, aiming to minimize all local objective functions simultaneously. This ensures that the global model resides on the Pareto front of the local loss functions, promoting fairness among clients. Every solution on the Pareto front satisfies a fairness criterion, as no local loss function can be improved without degrading at least one of the others.

To solve this MoM problem, each communication round begins with clients transmitting their local loss functions~to the PS. The PS then computes an adaptive weighting coefficient for gradient aggregation using a \textit{modified} version of the Chebyshev approach \cite{collette2013multiobjective}, a well-established technique in MoM. To facilitate gradient aggregation based on these adaptive weighting coefficients via the MAC channel, we design optimized transmit scalars for the clients and a de-noising receive scalar for the PS. This ensures that when clients simultaneously transmit their gradients, the distortion between the aggregated signal and the desired weighted gradients is minimized.

In summary, the key contributions of this letter are as follows:
($i$) to the best of the authors' knowledge, OTA-FFL is the first fair FL algorithm compatible with OTA computation principles;
($ii$) within the OTA-FFL framework, we propose a \textit{modified} version of the Chebyshev approach that enables the PS to compute adaptive weighting coefficients for each client’s gradient;
($iii$) we derive analytical solutions for the optimal transmit scalars at the clients and the denoising scalar at the PS, ensuring minimal distortion in gradient aggregation;
($iv$) we validate the effectiveness of the proposed OTA-FFL algorithm through extensive performance evaluations.

\par{Notation:}
The operators $(\cdot)^{\mathsf{T}}$ and $(\cdot)^{\mathsf{H}}$ denote the transpose, Hermitian transpose, respectively. Symbol $[K]$ denotes the set of integers $\{1,2,\cdots,K\}$, $\{f_k\}_{k \in [K]}=\{f_1,f_2,\dots,f_K\}$ for a scalar/function $f$. Let $\bold{u}[d]$ denote the $d$-th element of vector $\boldsymbol{\bold{u}}$.
For two vectors $\boldsymbol{\bold{u}}, \boldsymbol{\bold{v}} \in \mathbb{R}^D$, we say $\boldsymbol{\bold{u}} \leq \boldsymbol{\bold{v}}$ iff $\bold{u}[d] \leq \bold{v}[d]$ for $\forall d \in [D]$, i.e., two vectors are compared w.r.t. partial ordering. The $K$ dimensional probability simplex is denoted by $\Delta^K$ and $\mathbb{E}(\cdot)$ represents the mathematical expectation.

\section{Preliminaries} 
To incorporate the concept of \textit{fairness} into  the OTA-FL framework, we invoke notions from \textit{multi-objective optimization}. This section provides some preliminaries.

\subsection{Multi-Objective Optimization}
In MoM, multiple objective functions are to be minimized simultaneously. These  functions can be conflicting or incompatible. Specifically, 
let $\boldsymbol{f}(\boldsymbol{\theta}) = [f_1(\boldsymbol{\theta}), \ldots, f_K(\boldsymbol{\theta})]^{\mathsf{T}}$ be a set of $K$ scalar objective functions defined over the same domain $\Theta$. The goal in MoM is to find an optimal point $\boldsymbol{\theta}^*$, when the minimization is defined with respect to \emph{partial ordering}. This point is called \textit{Pareto-optimal}, which we define in the sequel.

\begin{definition}[Pareto-optimal]
The point $\boldsymbol{\theta}^*$ is called Pareto-optimal and is represented as
\begin{align} \label{eq:minpareto}
 \boldsymbol{\theta}^* = \argmin_{\boldsymbol{\theta} \in \Theta} \boldsymbol{f}(\boldsymbol{\theta}),
\end{align}
if there exists \textit{no} $\boldsymbol{\theta} \neq \boldsymbol{\theta}^*$, such that $f_k(\boldsymbol{\theta} ) < f_k(\boldsymbol{\theta}^*)$ for an integer $k\in [K]$, and $f_j(\boldsymbol{\theta} ) \leq f_j(\boldsymbol{\theta}^*)$ for all $j\neq k$. In other words, $\boldsymbol{\theta}^*$ is Pareto-optimal if it is impossible to decrease any $f_k$ without increasing at least one other objective $f_j$ for $j\neq k$. 
\end{definition}
Pareto-optimal solutions form the \textit{Pareto front}, the set of function values at all such points. A weaker notion, \textit{weak Pareto-optimality}, is satisfied if no \( \boldsymbol{\theta} \) strictly improves all objectives, i.e., \( \boldsymbol{f}(\boldsymbol{\theta}) < \boldsymbol{f}(\boldsymbol{\theta}^*) \). In addition, Pareto-optimality relates to stationary points in MoM, termed \textit{Pareto-stationary}:

\begin{definition} \label{stationary}
(Pareto-stationary): A point $\boldsymbol{\theta}^*$ is Pareto-stationary if a convex combination of objective gradients at $\boldsymbol{\theta}^*$ equals zero, i.e., if weights $\lambda_1, \ldots, \lambda_K \geq 0$ adding to one exist such that $\sum_{k =1}^K \lambda_k \nabla f_k(\boldsymbol{\theta}^*) = 0$.
\end{definition}
The following lemma draws the connection between Pareto-stationary and optimality.

\begin{lemma}[\cite{mukai1980algorithms}] \label{lem:pareto} 
Any Pareto-optimal solution to \eqref{eq:minpareto} is Pareto-stationary. Conversely, if $f_1,\ldots,f_K$ are convex any Pareto-stationary solution is weakly Pareto-optimal.
\end{lemma}

\subsection{Scalarization and Chebyshev Method}
Lemma~\eqref{lem:pareto} suggests to search for a Pareto-optimal solution within the set of Pareto-stationary points. Several approaches have thus been developed to identify this set. In the sequel,~we discuss two popular approaches: namely, the \textit{linear scalarization} and \textit{Chebyshev} method.

\noindent $\bullet$ \textbf{Scalarization:}  
Scalarization simplifies multi-objective optimization (MoM) by reformulating it as a single-objective minimization (SoM) problem. Specifically, given \( \boldsymbol{\lambda} \in \Delta^K \) for some \( \Delta \subseteq \mathbb{R} \), the MoM in \eqref{eq:minpareto} is transformed into the SoM:  
\begin{equation} \label{eq:fedavg}
f_{\mathrm{SoM}}(\boldsymbol{\theta}) = \sum_{k=1}^K \lambda_k f_k(\boldsymbol{\theta}),
\end{equation}
where \( \boldsymbol{\lambda} \) represents the weights assigned to each objective. It can be shown that for certain choices of \( \boldsymbol{\lambda} \), the stationary point of \( f_{\mathrm{SoM}} \) corresponds to a Pareto-stationary solution of \eqref{eq:minpareto}. In practice, \( \boldsymbol{\lambda} \) is selected based on some prior knowledge.

\noindent $\bullet$ \textbf{Chebyshev Method:} 
Scalarization can only converge to Pareto-optimal points that lie on the \emph{convex} envelop of the Pareto front \cite{boyd2004convex}. However, in many problems the Pareto front is non-convex, and hence points identified via scalarization are not Pareto-optimal. An alternative approach which can also converge to non-convex points on the Pareto front is the Chebyshev method. This method focuses on the maximum deviation rather than forming a linear combination of objectives.

To find a point on the Pareto-front, the Chebyshev method solves the following minimax problem.  
\begin{align} \label{eq:cheb}
\min_{\boldsymbol{\theta} \in \Theta} \max_{\boldsymbol{\lambda} \in \Delta^K} \boldsymbol{\lambda}^{\mathsf{T}} (\boldsymbol{f}(\boldsymbol{\theta}) - \boldsymbol{\zeta}),
\end{align}  
for some $\Delta\subseteq \mathbb{R}$ and a fixed $\boldsymbol{\zeta} \in \mathbb{R}^K$, which ideally serves as a lower bound for $\boldsymbol{f}$. Note that in this formulation, $\boldsymbol{\lambda}$ is optimized in the inner loop, and hence does not represent a linear combination of the objectives. This non-linearity enables the optimization algorithm to explore solutions that might not be attainable with linear scalarization.


\section{Problem Formulation}
Consider an FL setting with $K$ clients whose local datasets $\mathcal{D}_1, \ldots, \mathcal{D}_K$ are drawn from \textit{heterogeneous}, i.e., generally different, distributions. The clients collaborate with a PS to train a shared model with parameters $\boldsymbol{\theta}\in \mathbb{R}^d$. To this end, client $k$ determines its local empirical risk as 
\begin{align}\label{eq:f_k}
f_k \left(\boldsymbol{\theta}\right) \triangleq \frac{1}{|\mathcal{D}_k|} \sum_{(\bold{u}, v) \in \mathcal{D}_k}  \ell (\bold{u}, v \vert \boldsymbol{\theta} )    ,
\end{align}
for some loss function $\ell (\bold{u}, v \vert \boldsymbol{\theta})$ that measures the difference between the output of the learning model, parameterized with $\boldsymbol{\theta}$, to the input sample $\bold{u}$ and the ground-truth $v$. In iteration $t$, client $k$ starts from the actual global model $\boldsymbol{\theta}_t$ and trains it \textit{locally} by performing (multiple iterations of) a gradient-based algorithm on $f_k \left(\boldsymbol{\theta}\right)$ and computing the first-order information $\bold{g}_{t,k}\in \mathbb{R}^d$. It shares the information with the PS through its communication link. The PS receives an aggregation of the transmitted local parameters and process them into a global information $\bold{g}_{t}\in \mathbb{R}^d$. It then updates the global model to $\boldsymbol{\theta}_{t+1}$. 

The communication between the clients and the PS occurs over a wireless link, with aggregation performed directly over the air. The model for the communication links and the OTA computation scheme are specified in Section~\ref{sec:OTA-FFL}, where we present the proposed OTA-FFL scheme. 
Note that due to the use of OTA computation scheme, the PS has access \textit{only} to the aggregated local parameters $\bold{g}_{t,k}$, not the individual ones.

\label{sec:prelem}
\subsection{Notion of Fairness}
Despite using shared parameters, clients may experience varying learning performance due to heterogeneity, leading to \textit{unfairness} \cite{li2019fair}. We quantify fairness using a standard metric commonly adopted in the FL literature \cite{li2019fair}.

\begin{definition}[Fairness metric] \label{def:fair}
  A model $\boldsymbol{\theta}_1$ is said to be 
fairer than $\boldsymbol{\theta}_2$ if the test performance distribution of $\boldsymbol{\theta}_1$
across the network is more uniform than that of $\boldsymbol{\theta}_2$, i.e.,
\begin{align}
    \mathsf{std}\left[ f_1(\boldsymbol{\theta}_1), \ldots, f_K(\boldsymbol{\theta}_1)\right] < \mathsf{std}\left[ f_1(\boldsymbol{\theta}_2), \ldots, f_K(\boldsymbol{\theta}_2)\right],
\end{align}
where $\mathsf{std}\left[\cdot\right]$ denotes the standard deviation.   
\end{definition}

Definition~\ref{def:fair} helps us formulate the objective of \textit{fair FL} (FFL): the aim is to develop a distributed training algorithm, which converges to the \textit{fairest} model.

\subsection{Classical Approaches to Fair FL via MoM}
The FFL framework can be observed as a multi-objective optimization: we intend to find a shared model $\boldsymbol{\theta}$ that is optimal for all local risks. This problem mathematically reduces to the MoM \eqref{eq:minpareto} whose objective $f_k$ is the empirical risk in \eqref{eq:f_k}, and the \textit{optimal} FFL design is given by a Pareto-optimal solution. Nevertheless, among all Pareto-optimal models, we look for the one that is fairest in the sense defined in Definition~\ref{def:fair}.  

Classical FL algorithms estimate the Pareto-optimal model using different MoM techniques. We briefly discuss the two important ones that are related to our proposed scheme.

\noindent $\bullet$ \textbf{FedAvg:}
The conventional FedAvg suggests to use linear scalarization technique: the global model is found by estimating the solution of the SoM in \eqref{eq:fedavg} with the weights 
\begin{align}\label{eq:lam_k}
\lambda_k^{\mathsf{avg}} = \frac{\vert\mathcal{D}_k\vert}{\vert\bigcup_{k = 1}^{K} \mathcal{D}_k\vert},
\end{align}
ensuring that they are proportional to the sizes of local datasets \cite{mcmahan2017communication}. As discussed in preliminaries, this approach 
can only converge to the Pareto-optimal points that lie on the \emph{convex} envelop of the Pareto-front. Nevertheless, in FL the Pareto-front is non-convex \cite{hamidi2024adafed}, implying that FedAvg cannot reach a Pareto-optimal global model. It thus trains an \textit{unfair} model where the accuracy differs across clients \cite{li2019fair,hamidi2024adafed}.

\noindent $\bullet$ \textbf{Agnostic FL:}
The agnostic FL (AFL) algorithm, proposed in \cite{mohri2019agnostic}, suggests to find the Pareto-optimal models via the Chebyshev method. 
More precisely, AFL finds the global model by solving the minmax problem in \eqref{eq:cheb} with local empirical risks as objectives, while setting $\boldsymbol{\zeta} = 0$.

\begin{remark}
Other FFL schemes, such as the multiple gradient descent algorithm (MoM) \cite{desideri2009multiple}, rely on perfect knowledge of individual local parameters \( \bold{g}_{t,k} \). However, this information is unavailable in our setting due to the communication-efficient aggregation scheme based on OTA computation; see Section~\ref{sec:OTA-FFL}.
\end{remark}

\section{Fair FL via Modified Chebyshev Scheme}
The naive application of the Chebyshev method to the FL problem can leas to an FFL scheme that compromises in terms of the average accuracy of the global model. In this section, we propose a modified version of the Chebyshev method that draws a trade-off between Pareto-optimality and average performance. We use this modified method to develop a new FFL scheme. 

\subsection{Modified Chebyshev Scheme} \label{sec:cheb}
A closer examination of the minmax problem in \eqref{eq:cheb} reveals that the Chebyshev method promotes fairness by prioritizing the minimization of the loss function for the worst-performing client. This one-sided focus on fairness can compromise the average performance. This is in contrast to FedAvg, whose primary goal is to optimize average performance. It is therefore reasonable to modify \eqref{eq:cheb} to establish a trade-off between fairness and average performance. To this end, we propose constraining the inner optimization of \eqref{eq:cheb} as
\begin{align}\label{eq:lambda_s}
\min_{\boldsymbol{\theta} \in \theta}  \left\lbrace \max_{\substack{\boldsymbol{\lambda} \in \Delta^K}}
 \boldsymbol{\lambda}^{\mathsf{T}}  (\boldsymbol{f}(\boldsymbol{\theta}) - \boldsymbol{\zeta}) \text{ s.t. }  \| \boldsymbol{\lambda} - \boldsymbol{\lambda}^{\mathsf{avg}} \|_{\infty}\leq \epsilon \right\rbrace,   
\end{align}
where $\boldsymbol{\lambda}^{\mathsf{avg}} \in \mathbb{R}^K$ whose $k$-th entry is $\lambda_k^{\mathsf{avg}}$ defined in \eqref{eq:lam_k}. It is easily seen that \eqref{eq:lambda_s} draws a trade-off between the FedAvg and AFL schemes: 
\begin{inparaenum}
    \item[($i$)] for $\epsilon = 0$, it reduces to FedAvg.
    \item[($ii$)] By setting $\epsilon = 1$, it recovers the Chebyshev approach, where $\boldsymbol{\lambda}$ is adjusted freely to achieve maximal fairness (note that $\lambda_k^{\mathsf{avg}}\leq 1$ for $k\in [K]$).
\end{inparaenum}
 In practice, $\epsilon \in (0, 1)$ allows for a balanced trade-off between these two potentially conflicting objectives.

\subsection{FFL Algorithm based on Modified Chebyshev Scheme} \label{sec:cheb-FL}
We next integrate the modified Chebyshev scheme into the FL framework by developing an iterative scheme that solves the problem in \eqref{eq:lambda_s} in a distributed fashion. Note that at this stage, we only develop the distributed optimizer and postpone its implementation in the wireless network to Section~\ref{sec:OTA-FFL}.

We propose a two-tier optimizer whose inner and outer tiers solve the inner maximization and outer minimization of \eqref{eq:lambda_s}, respectively. These tiers are illustrated in the sequel.

\noindent $\bullet$ \textbf{Inner Tier:}
At iteration $t$, the PS shares the current global model $ \boldsymbol{\theta}_{t} $ with all clients. Client $k$ then sends its local risk value $f_k(\boldsymbol{\theta}_t)$ to the PS. Note that this requires exchange of a \textit{scalar} parameter per \textit{FL iteration}. This is a negligible overhead that can be integrated in control signaling being performed periodically in the network.   
Upon receiving the risk values, the PS solves the following maximization problem
\begin{align}
\boldsymbol{\lambda}^{\star}_t
= 
\max_{\boldsymbol{\lambda} \in \Delta^K}
 \boldsymbol{\lambda}^{\mathsf{T}}  (\boldsymbol{f}(\boldsymbol{\theta}_t) - \boldsymbol{\zeta}) \text{ s.t. } \| \boldsymbol{\lambda} - \boldsymbol{\lambda}^{\mathsf{avg}} \|_{\infty}\leq \epsilon.
\end{align}
Note that both the constraint and feasible set are convex in this optimization problem. We can hence deploy projection onto convex sets (POCS) algorithm to efficiently find $\boldsymbol{\lambda}^{\star}_t$.

\noindent $\bullet$ \textbf{Outer Tier:}
Once $\boldsymbol{\lambda}^{\star}_t$ is determined, the outer tier solves the following minimization.
\begin{align} \label{eq:min}
\min_{\boldsymbol{\theta}} \sum_{k=1}^K
\lambda^{\star}_{t,k} (f_k(\boldsymbol{\theta}) - \zeta_k),   
\end{align}
for $\zeta_k$ given in \eqref{eq:lambda_s}. This minimization describes the same scalarization form that is solved by FedAvg. We hence invoke the standard approach based on distributed (stochastic) gradient descent (DSGD): each client computes its local gradient at $ \boldsymbol{\theta}_{t} $, i.e., it sets $ \bold{g}_{t,k} = \nabla \left( f_k(\boldsymbol{\theta}_t) - \zeta_k \right) $, and sends it to the PS. The PS \textit{ideally} aggregates these local gradients into their weighted sum whose weights are proportional to $ \boldsymbol{\lambda}^{\star}_{t} $, i.e., 
\begin{align} \label{eq:g_t}
\bold{g}_t = \sum_{k =1}^K \lambda^{\star}_{t,k} \bold{g}_{t,k}.
\end{align}
The PS then performs one step of gradient descent to update the global model parameter as $\boldsymbol{\theta}_{t+1} = \boldsymbol{\theta}_{t} - \eta_t \bold{g}_t$ for some global learning rate $ \eta_t > 0 $. 

The proposed FFL scheme iterates between the inner and outer tiers until convergence, allowing the model to be updated based on both fairness and average performance considerations. In general, the inner and outer tiers can be updated at different rates. Our numerical however shows that the proposed scheme performs efficiently even in its basic form.

\begin{remark}
Intuitively, the proposed FFL scheme is an ``adaptive" form of FedAvg: unlike FedAvg whose weighting $\boldsymbol{\lambda}^{\mathsf{avg}}$ remains constant throughout the training loop, it uses $\boldsymbol{\lambda}^{\star}_t$ in iteration $t$, scheduled via the modified Chebyshev method.
\end{remark}

\section{Extension to OTA-FFL Scheme} \label{sec:OTA-FFL}
We extend the proposed scheme to the underlying wireless setting. In this respect, we develop an analog computation and scheduling scheme to aggregate gradients over the air. This is crucial, as it can reduce communication overhead significantly. 

\subsection{Communication Model and Over-the-Air Computation}
The clients are connected to the PS through a fading Gaussian MAC. Without loss of generality, we assume that clients and PS are equipped with a single antenna, and that at the beginning of iteration $t$, a subset of clients, denoted by $\mathcal{S}_t \subseteq [K]$, are scheduled to share their information. Client $k$ computes $\bold{x}_{t,k} \in \mathbb{R}^d$ from $\bold{g}_{t,k}$, which satisfies the power constraint $\mathbb{E} [\vert\bold{x}_{t,k}[i]\vert^2]\leq \mathsf{P}_0$ for $i \in [d]$ and a positive $\mathsf{P}_0$. It then transmits the signal over the MAC. Assuming the channel remains constant during one FL iteration, the received signal after \( d \) transmissions, denoted by \( \bold{y}_t \in \mathbb{C}^d \), is given by  
\begin{align}\label{eq:channel}
    \bold{y}_t = \sum_{k\in \mathcal{S}_t } h_{t,k} \bold{x}_{t,k} + \bold{n}_t,
\end{align}
where $h_{t,k}\in\mathbb{C}$ is the channel coefficient between client $k$ and the PS in iteration $t$, and $\bold{n}_t$ is complex zero-mean white Gaussian noise with variance $\sigma^2$, i.e., $\bold{n}_t \sim \mathcal{CN}(\boldsymbol{0},\sigma^2\bold{I}_d)$.

To reduce overhead, we leverage OTA computation, which allows the PS to estimate the global information $\bold{g}_{t}$ directly from the superimposed received signal $\bold{y}_t$ without any need to allocate separate time or frequency resources to each client. The OTA computation scheme consists of two modules: $K$ \textit{encoders} with its $k$-th one at client $k$ constructing $\bold{x}_{t,k}$ from $\bold{g}_{t,k}$, and a \textit{decoder} at the PS that estimates $\bold{g}_{t}$ from $\bold{y}_t$. 



\subsection{OTA Encoding-Decoding Scheme}

Local gradient entries can vary drastically in magnitude. The clients hence need to normalize them for pre-equalization and power control; see \cite{yang2020federated}. To this end, let us focus on iteration $t$, where in its inner tier the PS computes $\boldsymbol{\lambda}_t^\star$. The selected client $ k \in \mathcal{S}_t $ normalizes its local gradient $ \bold{g}_{t,k} $ into a symbol vector $ \bold{s}_{t,k} \in \mathbb{C}^d $ as follows: it first computes the mean and variance of its local gradient, denoted by $ m_{t,k} $ and $ v_{t,k} $, respectively and shares it with the PS over the uplink control channels. Note that these statistics are computed independent of loss values $f_k(\boldsymbol{\theta}_t)$, used by the FFL scheme, and hence can be shared through same control link. The PS upon collecting the local statistics $ \{ (m_{t,k}, v_{t,k}) : {k \in \mathcal{S}_t} \} $ computes the global mean and variance by weighted averaging as
\begin{subequations}
  \begin{align}
m_t = \sum_{k \in \mathcal{S}^t} \lambda^{\star}_{t,k} m_{t,k},  \qquad
v_t = \sum_{k \in \mathcal{S}^t} \lambda^{\star}_{t,k} v_{t,k}.
\end{align}  
\end{subequations}
The PS broadcasts the global statistics $ (m_t, v_t) $, and client $ k $ normalizes its gradient as $\bold{s}_{t,k} = \frac{1}{\sqrt{v_t}} \left( \bold{g}_{t,k} - m_t \bold{1}_d \right)$. It is straightforward to show that $\bold{s}_{t,k}$ satisfies $\mathbb{E} \left[ \bold{s}_{t,k} \bold{s}_{t,\ell}^{\mathsf{H}} \right] = \bold{0}_d$ for $\ell \neq k$ and $\mathbb{E} \left[ \bold{s}_{t,k} \bold{s}_{t,\ell}^{\mathsf{H}} \right] = \bold{I}_d$ for $\ell = k$.

The selected clients use linear scaling to encode their symbol vectors: client $ k \in \mathcal{S}_t $ transmits $\bold{x}_{t,k} = b_{t,k} \bold{s}_{t,k}$ for some $b_{t,k}$ which satisfies the transmit power constraint, i.e., 
\begin{align} \label{eq:power}
\mathbb{E} [|b_{t,k} \bold{s}_{t,k}[i]|^2] = |b_{t,k}|^2 \leq \mathsf{P}_0,
\end{align}
for $i\in[d]$, over $d$ channel uses. 
%
The received signal vector at the PS after $d$ channel uses is hence given by
\begin{align}\label{eq:aggr}
{\bold{y}}_t = \sum_{k \in \mathcal{S}^t} h_{t,k} b_{t,k} \bold{s}_{t,k} + \bold{n}_t,   
\end{align}
for some noise process $\bold{n}_t$ as defined in \eqref{eq:channel}. 

The received vector in \eqref{eq:aggr} can be seen as a noisy aggregation of normalized local gradients. We can hence process it via a decoder to compute an estimator of \eqref{eq:g_t} directly without individually estimating each local gradient. This explains the idea of OTA computation. Consider the linear encoding by clients, %
the PS further uses a linear decoding: it applies a de-noising receive scalar $ c_t $ to the received signal to compute an estimator of the normalized global gradients, and then 
utilizes the global statistics to estimate $ {\bold{g}}_t $ in \eqref{eq:g_t} when computed over the selected clients. The estimator is given by
\begin{align}\label{eq:hat_g}
\hat{\bold{g}}_t = \sqrt{v_t} \frac{{\bold{y}}_t}{c_t}  + m_t \bold{1}_d 
&= \sum_{k \in \mathcal{S}^t} \frac{h_{t,k} b_{t,k}}{c_t} \bold{g}_{t,k} + \frac{\sqrt{v_t}}{c_t} \bold{n}_t.
\end{align}

We design the tunable design parameters, i.e., $b_{t,k}$ and $c_t$, such that $\hat{\bold{g}}_t $ computes an \textit{unbiased} estimator of $\bold{g}_t $ with minimal variance. For an unbiased estimator, $\hat{\bold{g}}_t$ needs to satisfy
\begin{align}
\mathbb{E} [ \hat{\bold{g}}_t ] = \bold{g}_t .\label{eq:bias}
\end{align}
Under this condition, the variance is given by 
\begin{align}
\mathcal{E} (\hat{\bold{g}}_t, \bold{g}_t)   = \mathbb{E} \Big[ \| \hat{\bold{g}}_t - \bold{g}_t \|_2^2\Big]. \label{eq:MSE}
\end{align}
The transmit scalars $\{ b_{t,k}\}$ and the de-noising receive scalar $c_t$ are hence designed, such that \eqref{eq:bias} is satisfied, and \eqref{eq:MSE} is minimized simultaneously. The optimal design under these constraints is given in the following lemma whose proof is differed to the Appendix. 
\begin{lemma} \label{eq:lemma}
For an unbiased estimator $\hat{\bold{g}}_t$, $\{ b_{t,k}\}$ and $c_t$ that minimize the estimation variance are given by
\begin{align}
b_{t,k} = \frac{\lambda^{\star}_{t,k} c_t}{h_{t,k}}, \qquad 
c_t = \min_{k \in \mathcal{S}^t} \frac{\sqrt{\mathsf{P}_0} |h_{t,k}|}{\lambda^{\star}_{t,k}}.
\end{align}
For these optimal choices, the estimation variance is 
\begin{align} \label{eq:MSE_opt}
\mathcal{E}^\star  = \frac{d v_t \sigma^2}{\mathsf{P}_0} \max_{k \in \mathcal{S}^t} \frac{(\lambda^{\star}_{t,k})^2}{|h_{t,k}|^2}.
\end{align}
\end{lemma}

\subsection{Client Scheduling} \label{sec:schedule}
The error in \eqref{eq:MSE_opt} depends on the selected devices, \( \mathcal{S}_t \), creating a trade-off: expanding \( \mathcal{S}_t \) increases OTA computation error but reduces the variance of \( \bold{g}_t \). Optimal scheduling of \( \mathcal{S}_t \) is NP-hard \cite{bereyhi2023device}, leading to sub-optimal methods like channel-based scheduling \cite{amiri2020machine} or sparse recovery \cite{bereyhi2023device}. We adopt the efficient Gibbs sampling method from \cite{9810113}, omitting details for brevity.

\section{Numerical Experiments}
We next validate the OTA-FFL algorithm through experiments benchmarking it against state-of-the-art methods\footnote{Code is available at \url{https://github.com/Shayanmohajer/OTA-FFL/tree/main}.}.

\subsection{Settings and Baselines}
\noindent $\bullet$ \textbf{Benchmarks:} We consider ($i$) \textit{OTA-FedAvg} which is a straightforward integration of FedAvg into OTA computation,  
($ii$) \textit{OTA-TERM}, which is an adaptation of the TERM algorithm, whose local losses $f_k(\cdot)$ are replaced with $\mathsf{exp}\{\gamma f_k(\cdot)\}$ \cite{li2020tilted}, and 
($iii$) \textit{OTA-q-FFL} which is an adaptation of the $q$-FFL algorithm that replaces the local losses with $q^{\gamma f_k(\cdot)}$ \cite{li2019fair}. 


\noindent $\bullet$ \textbf{Datasets and setups:} We consider four datasets, with the following experimental configurations: for ($i$) \textit{ CIFAR-10}, we distribute the data non-iid among $K=10$ clients using Dirichlet allocation with $\beta=0.5$. As the model, we use ResNet-18 with Group Normalization and perform 100 communication rounds with all clients participating. We further set batch-size is set to $64$, consider a single local epoch at clients ($e=1$), and $\eta_t=0.01$. %
%
%
For ($ii$) \textit{CINIC-10}, use the same approach to distribute data non-iid among $K=50$ clients, and train the same model with the same parameters over 200 rounds with all clients participating. With ($iii$) \textit{FEMNIST}, we use a CNN with 2 convolutional layers followed by 2 fully-connected layers. The batch size is 32. We randomly sample $K=500$ clients from total of $K=3550$ and train the model using the default data stored in each client, with each client training for \textit{two local epochs}. For ($iv$) \textit{Fashion MNIST}, we use a fully-connected network
with 2 hidden layers, and use the same setting as that used in \cite{li2019fair}. We train the model with full batches for 300 rounds with a single local epoch and $\eta=0.1$.

\noindent $\bullet$ \textbf{Communication links:} Similar to \cite{10381881}, we consider channels with
10 distinct noise deviation values $\{0.1i: i\in[10]\}$, with the same number of channels for each type.

\noindent $\bullet$ \textbf{Performance metrics:} Let $ a_k $ represent the prediction accuracy of client $ k $. For each scheme, we compute the average test accuracy $\bar{a}$ (i.e., the mean of $\{a_1, \ldots, a_K\}$) and the standard deviation $\sigma_a$ \cite{li2019fair}. Also, similarly to \cite{li2020tilted}, we report the worst and best $10\%$ ($5\%$) accuracies.


\begin{table*}[!t]
\caption{\normalfont Fairness metrics evaluated on the test accuracy across  the CIFAR-10, CINIC-10, FEMNIST and Fashion MNIST datasets.}
\vskip -0.32in
\label{tableres}
\begin{center}
\begin{small}
\resizebox{\textwidth}{!}{%
\begin{tabular}{l|cccc|cccc|cccc|cccc}
\toprule
 & \multicolumn{4}{|c|}{CIFAR-10} & \multicolumn{4}{|c|}{CINIC-10} & \multicolumn{4}{|c|}{FEMNIST} & \multicolumn{4}{|c}{Fashion MNIST}
\\
\midrule \midrule
Algorithm & $\bar{a}$ & $\sigma_a$ & Worst 10\%  & Best 10\% & $\bar{a}$ & $\sigma_a$  & Worst 10\%  & Best 10\% & $\bar{a}$ & $\sigma_a$  & Worst 10\%  & Best 10\% & $\bar{a}$ & $\sigma_a$  & Worst 10\%  & Best 10\% \\
\midrule 
OTA-FedAvg   & \underline{62.32} & 5.41 & \underline{52.11} & \underline{71.05} &  \textbf{86.57} &17.66 &57.70 & 99.80 & 79.33 & 11.55 & 57.82 & 89.15 & \textbf{80.42} & 3.39 & 73.21 & 86.12\\
OTA-TERM & 57.05 & \underline{4.88} & 49.69 & 66.77 &  84.40 & 15.10 & 56.30 & \textbf{100.00} & \textbf{80.02} & \textbf{10.32} & \underline{61.37} & \textbf{90.16} & 79.29 & 2.53 & 74.36 & 85.21\\
OTA-q-FFL & 56.27 & 5.60 & 46.29 & 65.92 & 83.57 & \underline{14.91} &55.91 & 99.95 & 79.82 & 10.60 & 60.15 & 89.23 & 78.52 & \underline{2.27} & \underline{75.25} & \textbf{86.17}\\
OTA-FFL & \textbf{63.44} & \textbf{4.48} & \textbf{57.32} & \textbf{71.25} & \underline{86.41} & \textbf{13.80} & \textbf{57.89} & \underline{99.95} & \underline{80.01} & \textbf{7.12} & \textbf{65.85} & \underline{89.57} & \underline{79.59} & \textbf{2.12} & \textbf{76.28} & \underline{76.15} \\
\bottomrule
\end{tabular}}
\end{small}
\end{center}
\vskip -0.25in
\end{table*}

\subsection{Numerical Results}
The results are shown in \cref{tableres} where \textbf{bold} and \underline{underlined} numbers denote the best and second-best performance, respectively. Results are averaged over 5 random seeds. These results show that the proposed OTA-FFL algorithm consistently trains a significantly \textit{fairer} model compared to benchmarks. Interestingly, in some cases, the average accuracy also improves due to OTA-FFL's ability to dynamically balance client contributions and foster robust optimization in heterogeneous settings. To further illustrate the variation in client accuracies, \cref{fig:dist} presents a histogram of local accuracies for 500 clients on the FEMNIST dataset. The OTA-FFL accuracy distribution is more concentrated, reflecting improved fairness. 

\begin{figure}[t] \vspace{-.2cm}
	\centering
{\includegraphics[width=0.9\columnwidth,height=2.7cm]{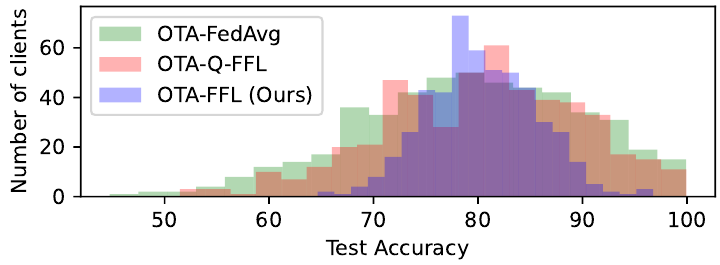}\label{fig:dist}}
\vskip -0.17in
\caption{The distribution of local accuracies for FEMNIST dataset.}
	\label{fig:dist}
\vskip -0.2in
\end{figure}

\noindent $\bullet$ \textbf{Computational Complexity:}
The additional computation required by OTA-FFL compared to OTA-FedAvg is to solve the optimization problem in \eqref{eq:lambda_s} using the POCS method. This introduces negligible overhead. Therefore, the overall running time of OTA-FFL is comparable to that of OTA-FedAvg.

\section{Conclusion}  
We have proposed a novel FFL algorithm based on the MoM formulation for FL, leveraging the Chebyshev method to minimize variation in local losses across clients. This approach ensures satisfactory performance for all clients. Extending to a wireless setting, we have integrated OTA computation, allowing the PS to aggregate local models directly over the air. Numerical results validate the effectiveness of our approach.

\appendix
\section{Appendix}
Starting from \eqref{eq:hat_g}, it is readily seen that \eqref{eq:bias} is satisfied when $b_{t,k} = \lambda^{\star}_{t,k} c_t / h_{t,k}$ for $k \in \mathcal{S}_t$. Replacing in \eqref{eq:MSE}, we conclude that $\mathcal{E} (\hat{\bold{g}}_t, \bold{g}_t)   = {d v_t}/{c_t^2}$ using the distribution of $\bold{n}_t$.

%
To guarantee the satisfaction of the transmit power constraint in \eqref{eq:power}, we need to have $|b_{t,k}^2| = \frac{(\lambda^{\star}_{t,k})^2 |c_t|^2}{|h_{t,k}|^2} \leq  \mathsf{P}_0$. Consequently, $\vert c_t \vert^2 \leq {\mathsf{P}_0 |h_{t,k}|^2}/{(\lambda^{\star}_{t,k})^2}$ or equivalently
\begin{align}
c_t \leq \min_{k \in \mathcal{S}^t} \frac{\sqrt{\mathsf{P}_0} |h_{t,k}|}{\lambda^{\star}_{t,k}}.
\end{align}
This concludes the proof. 
$\hfill \Box $

\bibliography{main}
\bibliographystyle{IEEEtran}

\end{document}